\documentclass[11pt]{article}  
\usepackage[margin=1in]{geometry}  
\usepackage{amsmath,amssymb,amsfonts}  
\usepackage{graphicx}  
\usepackage{booktabs}  
\usepackage{natbib}  
\usepackage{url}  
\usepackage{hyperref}  
\hypersetup{colorlinks=true,linkcolor=blue,citecolor=blue,urlcolor=blue}  
  
\title{Conformal Prediction Algorithms for Time Series Forecasting: Methods and Benchmarking}  
\author{Andro Sabashvili\thanks{Email: \href{mailto:andro.sabashvili@gmail.com}{andro.sabashvili@gmail.com}}}
\date{}  
  
\begin{document}  
\maketitle  
  
\begin{abstract}  
Reliable uncertainty quantification is of critical importance in time series forecasting, yet traditional methods often rely on restrictive distributional assumptions. Conformal prediction (CP) has emerged as a promising distribution-free framework for generating prediction intervals with rigorous theoretical guarantees. However, applying CP to sequential data presents a primary challenge: the temporal dependencies inherent in time series fundamentally violate the core assumption of data exchangeability, upon which standard CP guarantees are built. This paper critically examines the main categories of algorithmic solutions designed to address this conflict. We survey and benchmark methods that relax the exchangeability assumption, those that redefine the data unit to be a collection of independent time series, approaches that explicitly model the dynamics of the prediction residuals, and online learning algorithms that adapt to distribution shifts to maintain long-run coverage. We use AutoARIMA as the base forecaster on a large-scale monthly sales dataset, evaluating marginal coverage, interval width, and the Winkler score. Our benchmark results show that multi-step split conformal prediction method meets the 90\% coverage threshold and demonstrates the best efficiency.
\end{abstract}  
  
\section{Introduction}  
Point forecasts are ubiquitous, but modern decision-making in high-stakes domains such as energy, finance, and supply chains demands systematic uncertainty quantification. Prediction intervals, ranges expected to contain future outcomes with a specified probability, are essential for risk-aware planning. Conformal prediction (CP) is a powerful, model-agnostic framework that wraps around any base predictor (e.g., linear models, tree ensembles, neural networks) to deliver prediction sets with distribution-free guarantees \citep{vovk2005algorithmic,lei2018distribution}.  
  
A central conflict arises when applying CP to time series: standard guarantees rely on \emph{exchangeability}, the invariance of the joint distribution under permutation. Temporal dependence, autocorrelation, and nonstationarity violate exchangeability \citep{barber2023beyond}; na\"{i}ve application of CP then loses the coverage guarantees that make the framework appealing. Recent advances broaden conformal prediction beyond exchangeable settings by leveraging temporal dependence \citep{xu2021conformal, xu2023sequential, wang2024online}, online adaptation \citep{gibbs2021adaptive, bhatnagar2023online, angelopoulos2023pid}, weighted quantiles \citep{tibshirani2019covariate, barber2023beyond}, time-series datasets \citep{stankeviciute2021forecasting, sun2022copula}, and distribution shifts \citep{zou2024coverage}.

This study is practitioner-oriented and emphasizes simplicity, modularity, and speed. Accordingly, we use ARIMA as the base forecaster and deliberately exclude complex predictors such as neural network architectures. We further omit time series CP methods that are computationally intensive. Instead, we focus on plug-in conformal wrappers that can be layered on ARIMA with minimal overhead, are designed for the single-time-series setting, unlike the approaches in \citep{stankeviciute2021forecasting, sun2022copula}.

We structure the paper as follows. Section~\ref{sec:background} reviews split conformal prediction (SCP) and its guarantees. Section~\ref{sec:taxonomy} presents an overview of methods for time series: SCP adapted to multi-horizon forecasting, ensemble-based CP, CP that treats whole trajectories as exchangeable units, residual-dynamics modeling, and online adaptive controllers. Section~\ref{sec:bench} outlines experimental design, metrics commonly used to assess these methods and benchmark results. Section~\ref{sec:conclusion} summarizes the results of the study.
  
\section{Foundational Concepts}  
\label{sec:background}  
  
\subsection{Exchangeability}  
Exchangeability is the bedrock of standard conformal prediction guarantees. A sequence $\{Z_i\}_{i=1}^n$ (with $Z_i=(X_i,Y_i)$ for regression) is exchangeable if its joint distribution is invariant under permutations. The independent and identical distribution (i.i.d.) implies exchangeability, but exchangeability is strictly weaker. Under exchangeability, calibration scores computed on past data are representative of the score that would be observed for a new test point, enabling valid predictive inference without strong parametric assumptions \citep{shafer2008tutorial,lei2018distribution}.  
  
\subsection{SCP}  
SCP is a simple, widely used procedure for regression:  
\begin{enumerate}  
    \item \textbf{Split the data.} Partition the historical data into a proper training set and a calibration set.  
    \item \textbf{Fit a base model.} Train a predictor $f$ on the proper training set to produce point forecasts $\hat{y}$.  
    \item \textbf{Compute nonconformity scores.} For regression, the absolute residual $s_i = |y_i - \hat{y}_i|$ is common. Compute $\{s_i\}$ on the calibration set and sort them.  
    \item \textbf{Construct intervals.} For a new input $x_{\text{new}}$, predict $\hat{y}_{\text{new}}=f(x_{\text{new}})$ and set  
    $  
    \Gamma_{\alpha}(x_{\text{new}}) = \big[\, \hat{y}_{\text{new}} - q_{1-\alpha},\; \hat{y}_{\text{new}} + q_{1-\alpha} \,\big],  
    $  
    where $q_{1-\alpha}$ is the $\lceil (1-\alpha)(n_{\text{cal}}+1)\rceil$-th order statistic of the calibration residuals (finite-sample correction).  
\end{enumerate}  
  
\subsection{Validity Guarantee}  
Under exchangeability of the calibration and test points, SCP achieves \emph{marginal coverage}:  
$  
\mathbb{P}\big( Y \in \Gamma_{\alpha}(X) \big) \ge 1-\alpha,  
$  
which holds on average across test points. This differs from stronger \emph{conditional coverage} $\mathbb{P}(Y \in \Gamma_{\alpha}(X)\,|\,X=x)\ge 1-\alpha$, which is generally unachievable without further structure \citep{lei2018distribution,angelopoulos2023gentle}. Time series dependencies violate exchangeability, motivating adapted algorithms.  
  
\section{Conformal Prediction Algorithms for Time Series
}  
\label{sec:taxonomy}  
We divide methods by their strategy for handling non-exchangeability: (i) SCP adapted to horizon-specific calibration, (ii) relaxing assumptions via ensembles under mixing conditions, (iii) redefining the exchangeable unit from points to independent series, (iv) modeling residual dynamics for adaptive quantile estimation, and (v) online controllers for long-run coverage.  
  
\subsection{Multi-step Split Conformal Prediction (MSCP)} 
\textbf{Core principle.} MSCP adapts standard SCP to the multi-horizon forecasting setting by constructing \emph{horizon-specific} calibration scores and quantiles \citep{wang2024online, wang2026conformalForecast}. Unlike standard CP, which often relies on random splitting, time-series data requires a sequential split to preserve the underlying temporal dependencies and prevent data leakage.

\textbf{Methodology.} Let a base forecaster produce $\hat{y}_{t+h\mid t}$ for horizons $h=1,\dots,H$. Following a rolling window evaluation strategy, the model is initially fit on the proper training set to generate point forecasts and corresponding nonconformity scores $s_{t+h|t} = |y_{t+h} - \hat{y}_{t+h\mid t}|$. As new data points arrive, the training and calibration sets are continuously rolled forward, allowing the residual matrix $S\in\mathbb{R}^{T\times H}$ to be updated in an online learning framework. For each horizon $h$, we collect the residuals in the $h$-th column of $S$ and compute an empirical quantile $q_{h,1-\alpha}$. The horizon-specific prediction interval at time $t$ is:$$\Gamma_{\alpha}^{(h)}(t) = \big[\, \hat{y}_{t+h\mid t} - q_{h,1-\alpha},\; \hat{y}_{t+h\mid t} + q_{h,1-\alpha}\,\big].$$This rolling, horizon-specific approach acknowledges that uncertainty typically grows with $h$ and allows the intervals to adapt to local changes in the data distribution. While exchangeability is technically violated within a single series, this sequential split CP framework provides a principled way to assess long-run coverage behavior while maintaining the temporal integrity of the forecast.
  
\subsection{Ensemble Batch Prediction Intervals (EnbPI)}  
\textbf{Core principle.} EnbPI \citep{xu2021conformal}\footnote{Implementation of EnbPI: \url{https://github.com/hamrel-cxu/SPCI-code}}  relaxes the assumption of exchangeability by assuming the error process is stationary and \emph{strongly mixing}, allowing temporal dependence that decays sufficiently fast.  
  
\textbf{Methodology.} Train an ensemble of $B$ base models on bootstrap resamples of the full training set. For each training point $i$, form a leave-one-out (LOO) aggregate predictor $\tilde{y}_i$ by averaging only those ensemble members not trained on point $i$. The LOO residuals $\tilde{r}_i = |y_i - \tilde{y}_i|$ serve as calibration scores, leveraging all $T$ points without a strict train/calibration split. As data arrive, maintain a sliding window of residuals to adapt to recent dynamics.  
  
\textbf{Guarantee and assumptions.} Under stationary, strongly mixing errors and mild consistency of the base predictor, EnbPI yields approximate marginal coverage asymptotically. This trades finite-sample distribution-free validity for broader applicability to dependent data and improved sample efficiency via LOO calibration. 

\subsection{Sequential Predictive Conformal Inference (SPCI)}  
\textbf{Core principle.} Instead of enforcing exchangeability, treat residuals as a dependent signal to exploit. Model the dynamics of residuals and predict their quantiles directly \citep{xu2023sequential}\footnote{Implementation of SPCI: \url{https://github.com/hamrel-cxu/SPCI-code}}.
  
\textbf{Methodology.} Let $\{\hat{\varepsilon}_t\}$ be the sequence of realized residuals from the base forecaster. Fit a quantile regression model (e.g., quantile random forest, conformalized quantile regression) to predict future residual quantiles $(q_{\beta}, q_{1-\alpha+\beta})$ conditioned on lagged residuals. Construct intervals by shifting the point forecasts with the predicted residual quantiles: $\Gamma_{\alpha}(t) = [\,\hat{y}_{t+1\mid t} + q_{\beta},\;\hat{y}_{t+1\mid t} + q_{1-\alpha+\beta}\,]$, where $\beta$ minimizes the interval width.
  
\textbf{Guarantee and assumptions.} With consistent quantile estimation under dependence conditions, SPCI targets asymptotic validity closer to conditional coverage than purely marginal approaches, adapting to local dynamics. It avoids explicit distributional assumptions while capitalizing on serial correlation.

\subsection{Conformal Prediction for Time Series Dataset (Global-CP)}  
\textbf{Core principle.} When multiple independent time series are available, we treat each entire series as an exchangeable unit. This approach, adapted from the framework proposed by \citet{stankeviciute2021forecasting}, restores exchangeability at the \emph{trajectory} level rather than the time-step level.  
  
\textbf{Methodology.} For each series in the dataset, a local ARIMA model is trained to capture specific trend and seasonality. The dataset is split into training and calibration cohorts of series. After fitting local models, we compute absolute residuals on the held-out calibration cohort. For a multi-horizon forecast of length $H$, we pool these residuals across all calibration series to form horizon-specific distributions. While a standard conformal interval targets marginal validity at a single point, we aim to ensure the entire $H$-step trajectory remains within the bounds with probability $1-\alpha$. Following \citet{stankeviciute2021forecasting}, we apply a Bonferroni correction by setting the error budget for each horizon to $\alpha_h = \alpha/H$. The prediction interval at horizon $h$ is then constructed using the $(1 - \alpha/H)$ quantile of the pooled residuals:$$\Gamma_{\alpha}^{(h)} = [\hat{y}_{t+h} - q_{h, 1-\alpha/H}, \hat{y}_{t+h} + q_{h, 1-\alpha/H}]$$This ensures that the family-wise error rate is controlled across the full forecast trajectory. We evaluate this method specifically on its ability to provide joint coverage.
  
\textbf{Guarantee and assumptions.} This method is an adapted version of \citep{stankeviciute2021forecasting} which is based on a recurrent neural network to produce forecasts and provides finite-sample joint validity under exchangeability of series-level units. This paradigm is strong but more restrictive, presuming many similar, independent trajectories.  
  
\subsection{Online Learning and Control}  
\textbf{Core principle.} In streaming settings with distribution shift, treat coverage control as an online feedback problem: adjust interval parameters to meet a long-run coverage target regardless of shifts.  
  
\textbf{Adaptive Conformal Inference (ACI).} ACI \citep{gibbs2021adaptive, wang2026conformalForecast} iteratively updates the miscoverage level $\alpha_t$ based on recent coverage errors. Let $\mathrm{err}_t\in\{0,1\}$ indicate whether the interval at time $t$ failed to cover the true value. Update  
$  
\alpha_{t+1} \leftarrow \alpha_t + \gamma\,(\alpha - \mathrm{err}_t),  
$  
with learning rate $\gamma>0$. Intervals are constructed using the current $\alpha_t$ (e.g., via MSCP quantiles), enabling self-correction and convergence of average miscoverage to the target $\alpha$ under mild conditions.

\textbf{Autocorrelated Multi-step CP (AcMCP).} For multi-step forecasting, the $h$-step-ahead errors often exhibit MA$(h-1)$-like autocorrelation. AcMCP \citep{wang2024online, wang2026conformalForecast} incorporates an explicit forecast of future nonconformity scores into the controller. For horizon $h$, update the quantile estimate $q_{t+h\mid t}$ via  
\begin{align*}  
q_{t+h\mid t} \;&=\; q_{t+h-1\mid t-1} + \eta\,(\mathrm{err}_{t\mid t-h} - \alpha) \\  
&\quad + r_t\!\left( \sum_{i=h+1}^{t} (\mathrm{err}_{i\mid i-h} - \alpha) \right) + \tilde{e}_{t+h\mid t},  
\end{align*}  
where $\eta>0$ is a learning rate, $r_t(\cdot)$ is a saturation function to stabilize integral action, and $\tilde{e}_{t+h\mid t}$ is a forecast of the future nonconformity (e.g., via MA$(h-1)$ or linear regression on past errors). 
  
\textbf{Guarantee and assumptions.} Online controllers aim for \emph{long-run coverage}: the average miscoverage rate converges to $\alpha$. They are data distribution agnostic and are robust to nonstationarity, at the cost of stronger reliance on feedback tuning and potential efficiency loss during distribution shifts.  

\subsection{Standard Baselines}
In addition to the specialized algorithms described above, we evaluate parametric prediction intervals (Parametric-PI), which are derived analytically from the base forecaster’s forecast distribution. For ARIMA models, these typically assume Gaussian innovations, where the interval is defined by $\hat{y}_{t+h} \pm z_{1-\alpha/2} \hat{\sigma}_h$. While computationally efficient, this method lacks distribution-free guarantees and is sensitive to violations of the normality assumption. We utilize the implementation provided by Nixtla's \texttt{statsforecast} library \citep{garza2022statsforecast}. Furthermore, we evaluate the library’s native conformalized wrapper (Nixtla-CP), which employs a cross-validation procedure over multiple rolling historical windows to compute realized residuals. These residuals serve as nonconformity scores to calibrate the predictive range, providing a model-agnostic alternative to the purely parametric approach.

\section{Experimental Design and Benchmarking}  
\label{sec:bench}  
We compare the selected methods at the 90\% target coverage rate and evaluate interval validity and efficiency across diverse real-world datasets. We set a fixed forecasting horizon of $H=12$ months.  To ensure consistency, we employ the AutoARIMA model as the base forecaster for all experiments. We utilize the implementations from either Nixtla's \texttt{statsforecast} \citep{garza2022statsforecast} or the \texttt{forecast} \citep{hyndman2021fpp} libraries, depending on the method: the former is used for \texttt{EnbPI}, \texttt{SPCI}, \texttt{Parametric-PI}, \texttt{Nixtla-CP}, and \texttt{Global-CP}, while the latter supports the remaining adaptive and online methods. For all algorithms requiring parameter tuning, we maintain default hyperparameters to ensure a fair and reproducible benchmark.
    
\subsection{Dataset Description}  
A representative benchmark includes large-scale, real-world monthly sales data:  
\begin{itemize}  
    \item Scale: $>$3{,}000 individual time series.  
    \item Diversity: multiple countries and industries.  
    \item Sectors: entertainment, fashion, restaurant, electronics.  
\end{itemize}  
This heterogeneity tests robustness to varying seasonality, trends, and volatility.  
  
\subsection{Evaluation Metrics}  
To provide a comprehensive assessment, we evaluate the following metrics. For each metric, we first compute the value per series by averaging over all forecast horizons $h \in \{1, \dots, H\}$ in the test window. These series-level results are then averaged across the entire cohort of series to produce the final reported scores.\\ 
\textbf{Marginal coverage:} Empirical proportion of observations within intervals; target 90\% (i.e., $\alpha=0.1$). Validity requires empirical coverage $\ge 1-\alpha$.\\
\textbf{Joint coverage:} Evaluated specifically for the Global-CP method; defined as the proportion of test windows where the \emph{entire} $H$-step forecast trajectory is contained within the prediction intervals.\\
\textbf{Interval width:} Efficiency metric; among valid methods, narrower intervals are preferred.\\ 
\textbf{Winkler interval score (WIS):} Composite metric penalizing both miscoverage and unnecessary width. For observation $i$ with bounds $(\ell_i,u_i)$:
\begin{align*}  
\mathrm{WIS}_i =  
\begin{cases}  
(u_i - \ell_i) + \frac{2}{\alpha}(\ell_i - y_i), & \text{if } y_i < \ell_i, \\  
(u_i - \ell_i), & \text{if } \ell_i \le y_i \le u_i, \\  
(u_i - \ell_i) + \frac{2}{\alpha}(y_i - u_i), & \text{if } y_i > u_i.  
\end{cases}  
\end{align*}
Lower WIS is better.  

\subsection{Benchmark Results}
\begin{figure}
      \centering
      \includegraphics[width=0.7\linewidth]{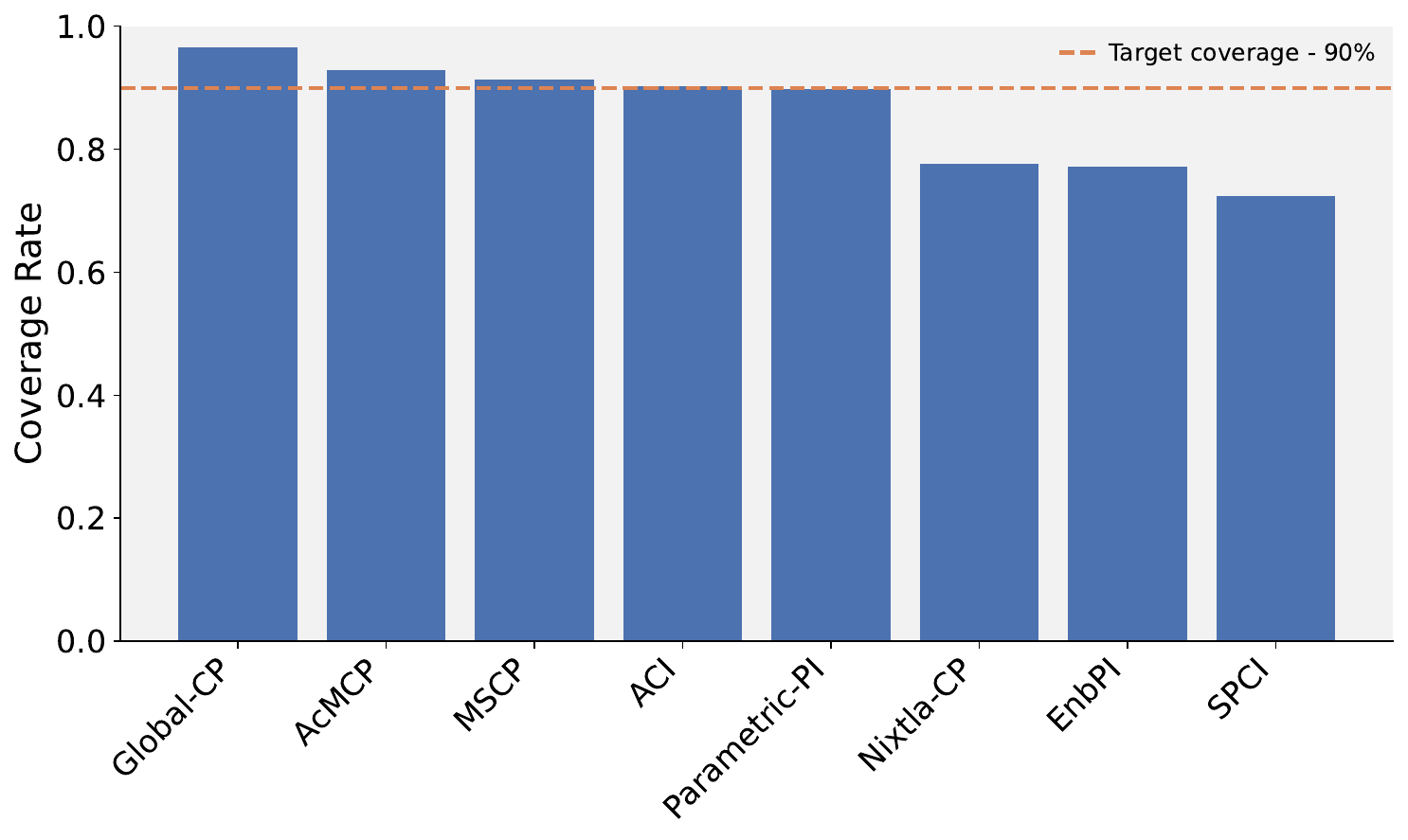}
      \caption{Empirical coverage rate (dashed orange line represents target coverage of 90\%).}
      \label{fig:coverage}
  \end{figure}

The benchmark results for the selected conformal prediction algorithms are evaluated based on their coverage validity and interval efficiency. The empirical coverage rates, illustrated in Fig.~\ref{fig:coverage}, reveal a distinct performance gap between the methods. Specifically, Global-CP, AcMCP, MSCP, ACI, and Parametric-PI successfully meet or exceed the 90\% target coverage threshold. In contrast, Nixtla-CP, EnbPI, and SPCI fail to provide the required coverage.

Efficiency is further examined through the Winkler interval score, which penalizes both excessive interval width and the magnitude of miscoverage.
To assess the statistical significance of the performance differences across the evaluated methods, we employ a two-step non-parametric procedure. First, the Friedman test is used to test the null hypothesis that all algorithms perform equally across the diverse datasets. Upon rejection of the null hypothesis ($p < 0.05$), we conduct a post-hoc analysis using the Conover-Friedman test \citep{conover1999practical} to perform pairwise comparisons. The results are visualized via a critical difference diagram, where average ranks are plotted and algorithms that do not show a statistically significant difference are connected by horizontal bars (cliques) based on the calculated critical difference.

\begin{figure}
    \centering
    \includegraphics[width=0.8\linewidth]{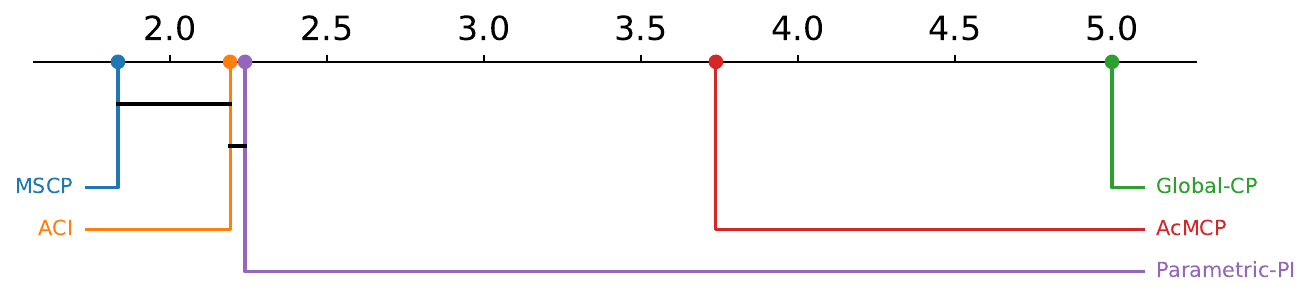}
    \caption{Critical difference diagram based on the Winkler interval score}
    \label{fig:winkler_cd}
\end{figure}

Initial testing indicates that MSCP, ACI, and Parametric-PI are the top three performing algorithms, Fig.~\ref{fig:winkler_cd}. While these methods significantly outperform approaches like Global-CP and AcMCP, the initial experimental data show no statistically significant difference between them. This suggests that for smaller datasets, all three provide comparable utility in terms of interval tightness and predictive accuracy.

\begin{figure}
    \centering
    \includegraphics[width=0.8\linewidth]{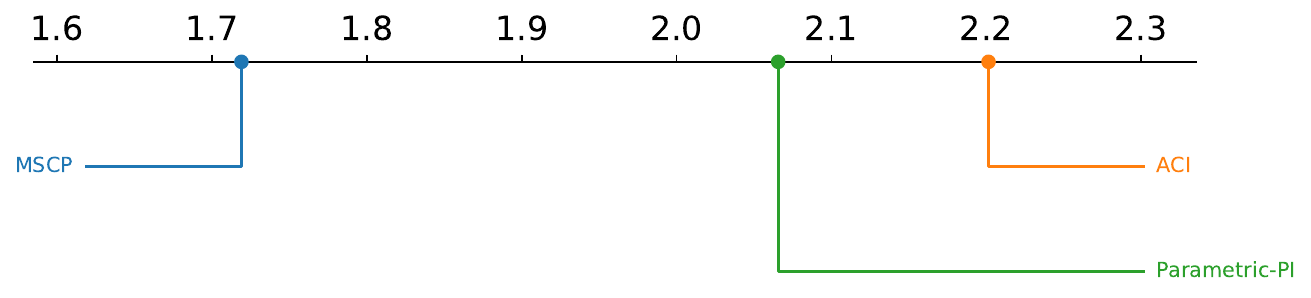}
    \caption{Critical difference diagram based on the Winkler interval score for the top 3 algorithms: MSCP, Parametric-PI, and ACI}
    \label{fig:final_winkler_cd}
\end{figure}

To resolve the ranking among the leading candidates, the analysis was extended to a larger corpus of time series data to enhance the power of the statistical significance tests. The finalized critical difference diagram for these top three algorithms is shown in Fig.~\ref{fig:final_winkler_cd}. With the increased number of observations, a clear hierarchy emerges: the MSCP method achieves the best result followed by Parametric-PI, while ACI ranks lowest among the three. These findings indicate that the MSCP method provides the most effective balance of coverage validity and interval efficiency for multi-horizon forecasting tasks.

\section{Conclusion}
\label{sec:conclusion}
The benchmarking of conformal prediction algorithms for time series forecasting reveals critical insights into the trade-off between theoretical coverage guarantees and practical interval efficiency. Our analysis confirms that while several methods, including Global-CP, AcMCP, MSCP, ACI, and Parametric-PI, successfully maintain the 90\% target coverage, there is a significant discrepancy in their empirical reliability. Notably, methods such as Nixtla-CP, EnbPI, and SPCI failed to reach the nominal coverage threshold in this experimental setting, with SPCI demonstrating the most pronounced under-coverage. This suggests that for ARIMA-based forecasting, relying on methods that assume high degrees of mixing or specific residual dynamics may lead to over-optimistic and ultimately invalid uncertainty estimates.

Beyond simple validity, the efficiency of the generated intervals—as quantified by the Winkler score—distinctly separates the valid algorithms. While initial testing showed that MSCP, ACI, and Parametric-PI were statistically comparable, the expansion of the benchmark to a larger corpus of time series data provided the necessary statistical power to resolve their performance hierarchy. The results demonstrate that the MSCP method, characterized by its horizon-specific calibration, achieves the highest efficiency. It is followed by Parametric-PI and ACI, which exhibited slightly higher average ranks, indicating wider or less calibrated intervals across the tested horizons.

In conclusion, this study highlights that for practitioners utilizing modular forecasting pipelines, the MSCP approach offers the most robust and efficient solution for distribution-free uncertainty quantification in multi-horizon forecasting. The success of this method emphasizes the importance of horizon-specific calibration in accounting for the temporal dependencies that typically violate the exchangeability assumption. While online controllers like ACI and AcMCP remain highly competitive and adaptable to non-stationary environments, the simplicity and superior efficiency of horizon-specific split conformal prediction make it the preferred choice for the high-stakes sequential tasks evaluated here.

\bibliographystyle{plainnat}

\end{document}